\documentclass[12pt]{article}

\usepackage{xcolor}
\usepackage{scicite}

\usepackage{times}

\newenvironment{sciabstract}{%
\begin{quote} \bf}
{\end{quote}}

\usepackage{booktabs}
\usepackage{amsmath}
\usepackage{amssymb}
\usepackage{setspace}
\usepackage{mathtools}
\usepackage{algpseudocode}
\usepackage{algorithm}
\usepackage{subcaption}
\usepackage{graphicx}
\usepackage{hyperref}
\usepackage{xcolor}
\usepackage{multirow} 

\DeclareMathOperator*{\argmin}{argmin} 
\DeclareMathOperator*{\argGN}{argGN} 

\title{Artificial Microsaccade Compensation:\\ Stable Vision for an Ornithopter}

\author
{Levi Burner,$^{1\ast}$ Guido C. H. E. de Croon,$^{2}$ Yiannis Aloimonos$^{1,3}$\vspace{10pt}\\
\normalsize{$^{1}$Department of Computer Science, University of Maryland, College Park}\\
\normalsize{$^{2}$Faculty of Aerospace Engineering, Delft University of Technology}\\
\normalsize{$^{3}$Institute for Advanced Computer Studies, University of Maryland, College Park}\\
\normalsize{$^\ast$Corresponding author. Email: lburner@umd.edu (Levi Burner)}
}

\date{}

\begin{document} 


\baselineskip24pt


\maketitle 


\begin{sciabstract}
\begin{center}
{\large\textbf{Abstract}}
\end{center}
Animals with foveated vision, including humans, experience microsaccades, small, rapid eye movements that they are not aware of. Inspired by this phenomenon, we develop a method for ``Artificial Microsaccade Compensation''. It can stabilize video captured by a tailless ornithopter that has resisted attempts to use camera-based sensing because it shakes at 12-20 Hz. Our approach minimizes changes in image intensity by optimizing over 3D rotation represented in $SO(3)$. This results in a stabilized video, computed in real time, suitable for human viewing, and free from distortion. When adapted to hold a fixed viewing orientation, up to occasional saccades, it can dramatically reduce inter-frame motion while also benefiting from an efficient recursive update. When compared to Adobe Premier Pro's warp stabilizer, which is widely regarded as the best commercial video stabilization software available, our method achieves higher quality results while also running in real time.
\end{sciabstract}


\newpage
\section*{Introduction}
Flapping flying robots, or ornithopters, stand to revolutionize micro air vehicles, because they are safer, quieter, resilient to contact, and more aesthetically appealing than their rotor based counterparts. However, their wing-based actuation induces shaking that makes camera-based perception difficult \cite{guido2020skills}. In particular, tailless flapping robots are of interest because they can hover and are more agile than their tailed counterparts \cite{Park2008-tailed-flapping, de2009design, Nian2020-tailed-flapping-dynamics, Phan2019-tailless-agility} and can even imitate agile insect behaviors \cite{karasek2018tailless}. However, tailless ornithopters also shake aggressively, which creates difficulties when using vision-based sensing \cite{tapia2023comparison, tapia2025flight, rafee2025review}. The deteriorated visual sensing is a major obstacle on the road to real-world application of flapping wing robots, as it makes it hard to use cameras for autonomous flight and applications.

A particularly popular platform is the Flapper Nimble+, which is a tailless ornithopter whose original design was used to demonstrate that fruit flies rely on torque coupling to perform rapid bank maneuvers \cite{karasek2018tailless}. As a result, the platform is quite agile, however, it also shakes aggressively because it cannot rely on the tail for damping.  Moreover, the rotations due to shaking are irregular since the two wing pairs are actuated independently to generate the moments necessary for flight maneuvers. There is potential for significant advancements in the mechanical capabilities of flapping robots \cite{lau2025spring}. However, applications remain limited because it has proven difficult for tailless flapping robots to control themselves via an onboard camera, given their limited payload capacity and aggressive vibration \cite{Olejnik2020}. 

One work introduced a mechanical camera stabilization system, however, this introduces complexity and weight, and so a software solution is preferred if possible \cite{pan2020stabilizer, rafee2025review}. Recent work has shown that event cameras are the ideal sensor for these platforms, however the weight of currently available systems, and their need for a powerful onboard computer has made implementation challenging \cite{tapia2023comparison, tapia2025flight}. Vision based control of a tailed flapping robot has been achieved with wireless monocular \cite{de2009design} and a specialized miniature stereo camera system \cite{de2014autonomous}.  However, rolling shutter effects remain a persistent challenge \cite{rafee2025review}. Recently, video stabilization has been shown to be beneficial for both frame and event based perception with a focus on flapping robots. However, the method relies on access to ground truth orientation \cite{rodriguez2024benefits}.  Hence, the shaking of vision sensors remains a challenging problem for autonomous flapping wing robots.

An interesting parallel can be drawn to animals with foveated vision, such as humans, who constantly experience rapid, small eye movements called microsaccades. These small rotational eye movements are not perceived in our consciousness, and while their purpose or lack thereof is still a matter of debate, there is a consensus that the brain employs several techniques to mask, or suppress, these movements so that the world appears to be viewed from a stable orientation \cite{poletti2016compact}. Thus, it is of interest to develop an artificial method for compensating microsaccade like motion for robots that shake.

In this paper, we describe a real-time method for video stabilization, which we call ``Artificial Microsaccade Compensation''. It directly minimizes differences in image intensity, without relying on feature matching. The resulting rotations are used to estimate a stable, slowly varying orientation to which frames are rendered via a stabilizing rotation. The optimization is conducted over the rotation group of $SO(3)$ using an efficient, equivariant approach based on the inverse compositional Lucas-Kanade method \cite{baker2004lucas}.

While direct methods for computing image motion have been known for some time \cite{lucas1981iterative}, modern open-source libraries lack the capabilities required to directly estimate nonlinear warps, such as 3D rotations, in a direct and efficient manner without resorting to feature matching. Consequently, we open-source our implementation, in the hope that other practitioners with shaking robots can stabilize their image streams by rendering them to a stable viewpoint.

\newpage
We make the following contributions:
\begin{itemize}
\item A framework and implementation for ``Artificial Microsaccade Compensation''. That makes clear how a robot can represent a video from a stabilized orientation  without apparent distortion, and which can be adapted for other applications.
\item An efficient implementation of a direct Lucas-Kanade style method for image rotation estimation based on equivariance and optimization in $SO(3)$ with updates computed in the Lie Algebra $so(3)$. To the best of our knowledge, an equivalent implementation is not available.
\item Evaluation of the method in a particularly difficult stabilization problem arising from the shaking of a tailless ornithopter, the Flapper Nimble+, which has thus far admitted only limited application of camera based sensing \cite{Olejnik2020}. The result is the first stable videos  from a camera on the platform.
\end{itemize}

\section*{Results}
\begin{figure}
  \centering
  \includegraphics[width=1.0\linewidth]{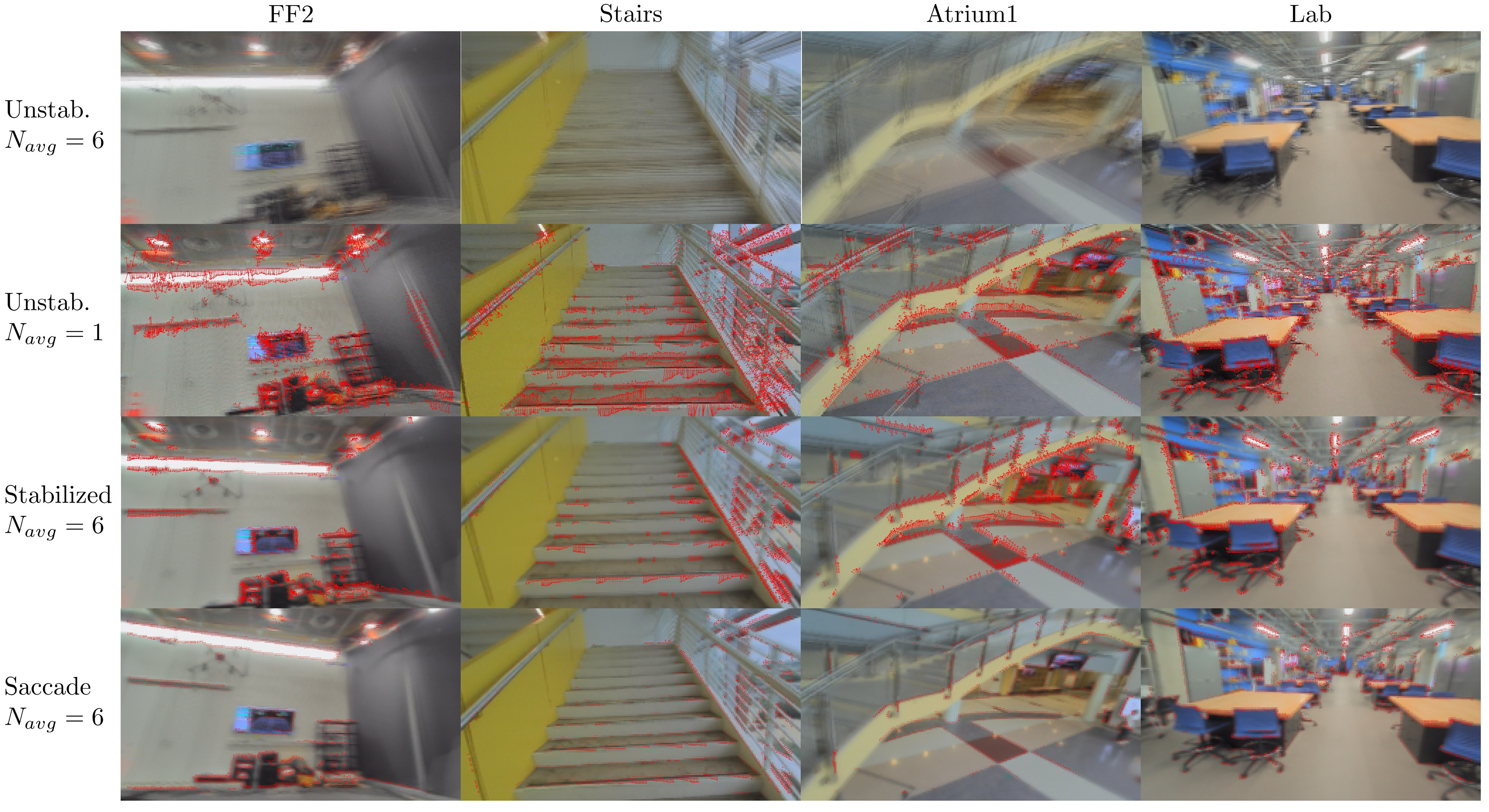}
  \caption{Qualitative comparison of stabilized and unstabilized frames. Top row, 6 averaged consecutive frames from a sequence illustrating the extreme shaking of a camera onboard the Flapper Nimble+ tailless ornithopter. Second row, a single unstabilized frame with normal flow (the projection of optical-flow onto gradients) illustrated with red arrows. Third row, 6 stabilized and averaged frames, which were warped to the stabilized viewpoint. Bottom row, 6 stabilized and averaged frames which were warped to a stationary viewpoint that occasionally sacccades. The flow magnitude can be seen to reduce with each row, with the saccading variation of the algorithm achieving a drastic reduction in flow magnitude. Videos showing stabilization results are available in the supplementary information. }
  \label{fig:qualitative}
\end{figure}

\begin{figure}
  \centering
  \includegraphics[width=0.5\linewidth]{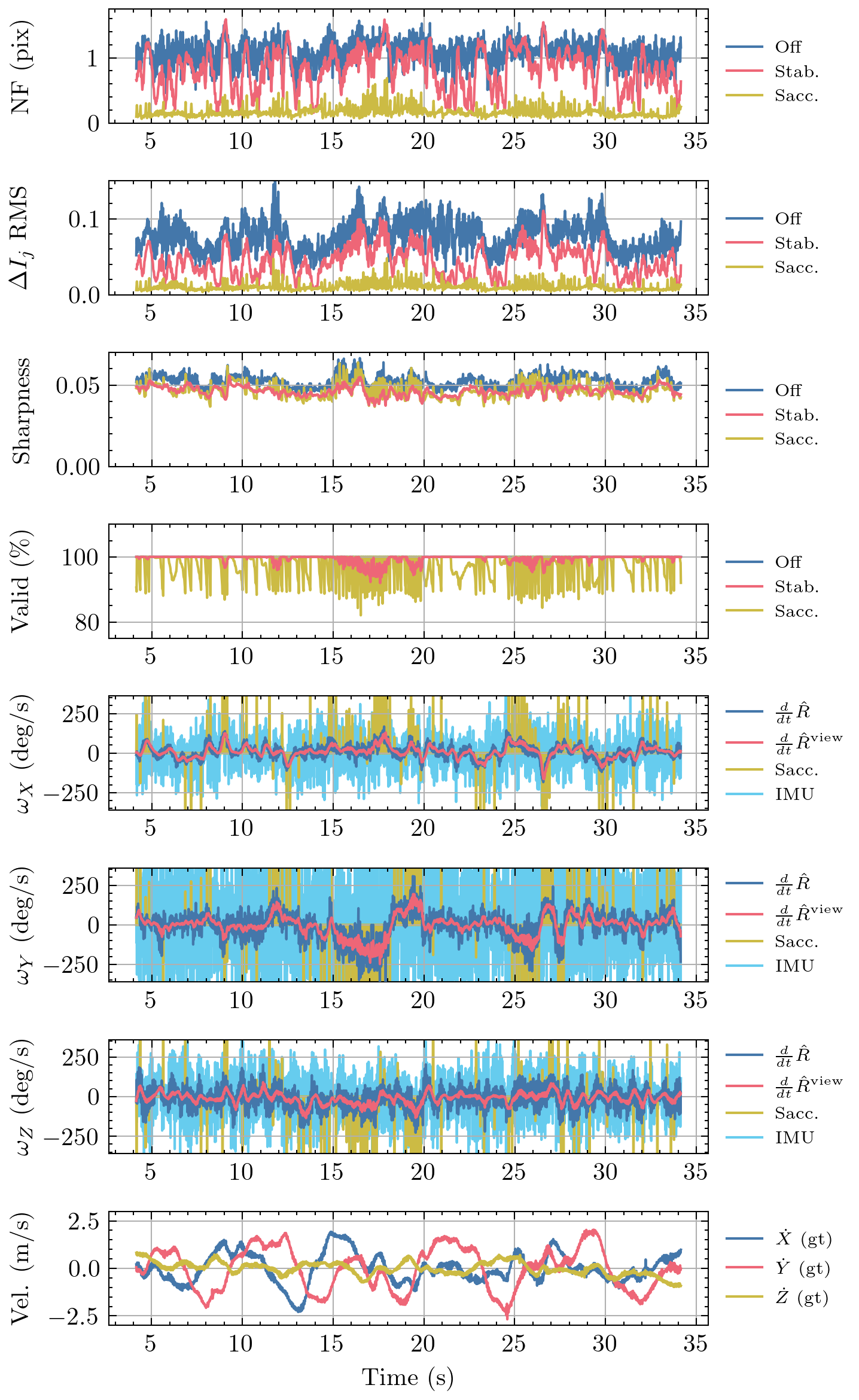}
  \caption{Per frame metrics computed on sequence FF2. Normal flow and the RMS change in image brightness are reduced by the stabilization and significantly reduced by saccading instead of continuously rotating the stable view. The proposed frame averaging, which reduces the effects of rolling shutter and image artifacts, slightly reduces the stabilized image sharpness. The stabilization algorithm maintains a high quantity of valid pixels given 12.5\% image margins. The saccading variant results in slightly fewer valid pixels. During flight, the angular velocities from the Flapper's onboard IMU do not align with the angular velocities of the image based orientation estimate. The stabilized orientation does not contain the high frequency oscillations of the estimated orientation. The saccade style stabilization maintains a viewpoint angular velocity of zero, except when it saccades, which results in an angular velocity spike.}
  \label{fig:qualitativeseq7}
\end{figure}

The algorithm was tested on a Flapper Nimble+ manufactured by Flapper Drones equipped with a wireless FPV camera. Videos showing stabilization results are available in the supplemental material. Figure \ref{fig:qualitative} shows several qualitative results, demonstrating a dramatic improvement of the image stability. In particular, the motion vectors from normal flow estimation, shown as red arrows, show that the apparent image motion is reduced drastically.
Figure \ref{fig:qualitativeseq7} shows per-frame metrics for sequence FF2. The metrics are described below.
Tables \ref{tab:mocap} and \ref{tab:outlab} detail quantitative results averaged over each sequence.

The videos in the supplementary information include side-by-side comparison with Adobe Premier Pro's warp stabilizer, which is widely regarded as the best commercial video stabilization software available. Adobe's documentation indicates that their method relies on feature matching and subspace warping \cite{adobestab}. In most sequences, the video stabilized by Adobe Premier Pro is still vigorously shaking, and frequent distortions are introduced which make nearby objects look like they are stretching. In some sequences, Adobe Premier Pro fails to produce a suitable result, with most of the frame remaining empty. The warp stabilizing plugin was configured with the maximum smoothness setting (100\%) and with defaults for all other settings. On some sequences, the software automatically advised disabling cropping, which we did. It must be noted that the Adobe software uses several additional gigabytes of RAM during stabilization, which is conducted in an offline, batch processing mode. Our method is capable of running online, achieves real-time performance, and uses several hundred megabytes total, most of which can be attributed to its research-quality code which is written in Python and uses the high level JIT compilation and autodifferentiation framework Jax \cite{jax2018github}. Specifically, autodifferentiation is used to precompute the static Jacobians used for every Gauss Newton step.

For all results, the images were downsampled by 4x to 320$\times$180 resolution. $12.5\%$ margins were applied to each side of the output frames to limit the appearance of border effects due to the stabilized view's field of view being outside of the real camera's field of view. The final output resolution was 240$\times$152. At this resolution, the algorithm achieved real-time performance as needed for future downstream applications. Specifically, on sequence Atrium2, it achieved 67 and 77 fps in the stabilized and saccade modes, respectively on a i7-12800HX laptop CPU. Note that our code is implemented in Python, is almost entirely single threaded, loads images from disk in an uncompressed format, and does not use hardware acceleration for image interpolation which would speedup undistortion, the Lucas-Kanade tracker, and computation of $I^{\mathrm{stab}}$. Figure \ref{fig:stabilization} contains examples of high-resolution stabilized frames without downsampling.

\begin{table*}
\centering
\resizebox{0.8\textwidth}{!}{%
\begin{tabular}{ll|rrrrrrrrrr}
\textbf{Metric} & \textbf{Method} &   \multicolumn{10}{l}{\textbf{Sequence}}  \\ \hline
            &          & UD    & LR    & FB    & Yaw   & Circle1 & Circle2 & FF1   & FF2   & FF3   & Hand  \\ \hline\hline
NF RMS & None & 1.047 & 1.106 & 1.077 & 1.168 & 1.082 & 1.085 & 1.120 & 1.102 & 1.155 & 1.025 \\
 (pix.)& Stab. & 0.598 & 0.739 & 0.801 & 0.959 & 0.855 & 0.810 & 0.874 & 0.876 & 0.973 & 0.912 \\
       & Sacc. & 0.129 & 0.155 & 0.157 & 0.207 & 0.169 & 0.175 & 0.182 & 0.174 & 0.197 & 0.103 \\ \hline
$\Delta I_j$ RMS & None & 0.075 & 0.080 & 0.070 & 0.089 & 0.073 & 0.076 & 0.083 & 0.079 & 0.080 & 0.078 \\
          & Stab. & 0.032 & 0.041 & 0.038 & 0.062 & 0.049 & 0.040 & 0.049 & 0.048 & 0.053 & 0.062 \\
          & Sacc. & 0.009 & 0.010 & 0.010 & 0.014 & 0.011 & 0.011 & 0.012 & 0.011 & 0.012 & 0.007 \\ \hline
Sharpness & None & 0.056 & 0.055 & 0.050 & 0.054 & 0.052 & 0.052 & 0.053 & 0.053 & 0.051 & 0.056 \\
          & Stab. & 0.050 & 0.049 & 0.045 & 0.046 & 0.046 & 0.046 & 0.047 & 0.046 & 0.045 & 0.051 \\
          & Sacc. & 0.050 & 0.048 & 0.045 & 0.048 & 0.047 & 0.045 & 0.047 & 0.047 & 0.046 & 0.051 \\ \hline
Valid Pix. & None & 100.0 & 100.0 & 100.0 & 100.0 & 100.0 & 100.0 & 100.0 & 100.0 & 100.0 & 100.0 \\
(\%)       & Stab & 100.0 & 99.8 & 99.7 & 97.1 & 99.2 & 99.9 & 99.4 & 99.4 & 99.1 & 99.2 \\
           & Sacc. & 97.8 & 97.4 & 97.6 & 96.8 & 97.2 & 97.4 & 97.2 & 97.1 & 97.2 & 96.7 \\ \hline
$\omega$ RMS & MoCap & 238 & 229 & 259 & 294 & 282 & 258 & 260 & 249 & 310 & 143 \\
(deg/s)      & IMU & 301 & 289 & 268 & 368 & 347 & 276 & 308 & 311 & 325 &  90 \\
                     & Image &  85 &  97 &  88 & 157 & 128 &  93 & 107 & 105 & 124 &  92 \\ \hline
$\omega^{\mathrm{view}}$ RMS & Stab. &  32 &  50 &  43 & 130 &  99 &  48 &  71 &  70 &  87 &  82 \\
(deg/s)                      & Sacc. &  21 &  44 &  39 & 129 &  97 &  42 &  66 &  65 &  89 &  75 \\ \hline
V RMS (m/s) & MoCap & 0.908 & 1.157 & 1.294 & 0.815 & 1.410 & 1.576 & 1.338 & 1.313 & 1.658 & 0.591 \\ \hline \hline
\end{tabular}%
}
\caption{Image quality, stabilization quality, angular velocity, and translational velocity metrics for ten sequences collected in a motion capture lab, each featuring different types of motion.}
\label{tab:mocap}
\end{table*}

\begin{table}
\center
\resizebox{0.5\columnwidth}{!}{%
\begin{tabular}{ll|rrrrr}
\textbf{Metric} & \textbf{Method} &   \multicolumn{5}{l}{\textbf{Sequence}}  \\ \hline
            &          & Hall    & Stairs    & Atrium1    & Atrium2   & Lab  \\ \hline\hline
NF RMS & None & 1.026 & 1.104 & 1.053 & 1.119 & 1.052 \\
(pix.)          & Stab. & 0.554 & 0.694 & 0.734 & 0.853 & 0.574 \\
              & Sacc. & 0.229 & 0.231 & 0.170 & 0.192 & 0.203 \\\hline
$\Delta I_j$ RMS & None & 0.048 & 0.073 & 0.078 & 0.079 & 0.078 \\
          & Stab. & 0.018 & 0.029 & 0.043 & 0.048 & 0.033 \\
          & Sacc. & 0.009 & 0.011 & 0.010 & 0.010 & 0.012 \\\hline
Sharpness & None & 0.047 & 0.053 & 0.054 & 0.053 & 0.060 \\
          & Stab. & 0.042 & 0.045 & 0.047 & 0.046 & 0.053 \\
          & Sacc. & 0.041 & 0.045 & 0.047 & 0.046 & 0.052 \\\hline
Valid Pix. & None & 100.0 & 100.0 & 100.0 & 100.0 & 100.0 \\
(\%)       & Stab. & 100.0 & 100.0 & 99.7 & 99.5 & 100.0 \\
           & Sacc. & 97.3 & 98.0 & 97.7 & 97.4 & 97.6 \\\hline
$\omega$ RMS & IMU & 250 & 273 & 273 & 276 & 234 \\
(deg/s)      & Image &  80 &  99 & 101 & 110 &  80 \\\hline
$\omega^{\mathrm{view}}$ RMS & Stab. &  24 &  39 &  57 &  63 &  32 \\
(deg/s)                      & Sacc. &  17 &  32 &  49 &  58 &  22 \\\hline\hline
\end{tabular}%
}
\caption{Image quality, stabilization quality, and angular velocity metrics for five sequences collected indoors.}
\label{tab:outlab}
\end{table}

\subsection*{Sequences}

Ten sequences, each approximately 20 seconds in length, were collected in a motion capture room. The Flapper was piloted manually to achieve the following trajectories. Up-down (UD), left-right (LR), front-back (FB), yaw in place, fly in a circle (Circle1, Circle2), fly freely according to the operators discretion (FF1, FF2, FF3). The tenth sequence was collected by holding the flapper in hand, and quickly tilting it back and forth along each primary axis, followed by a smooth figure 8 motion. This in-hand trajectory is meant to demonstrate the best possible image quality when the camera's motion is smooth.

An additional five sequences, each approximately 20 seconds in length, were collected in a variety of indoor settings. Due to airspace restrictions, we were unable to collect data outside. The sequences include a hallway, stairwell, a large atrium (2x), and a robotics laboratory. These sequences test the algorithm with a variety of visual features and scene depths.

\subsection*{Metrics}

We compute a variety of metrics to demonstrate the image stabilization quality. The first is the RMS difference between consecutive output images. As can be expected, the stabilized frames feature a noticeable reduction in RMS difference between subsequent frames, often achieving an approximately 50\% reduction. Further, the saccade variant, which results in stabilized frames with a fixed orientation, achieves an approximately 7$\times$ reduction. These results can be expected because the Lucas-Kanade tracker minimizes the MSE between frames, though not necessarily pairs adjacent in time.

Next, we estimate image motion in the unstabilized, stabilized, and saccade stabilized frames using RMS normal flow magnitude. Normal flow has units of pixel displacement that is the magnitude of the optical flow onto the image gradients and can be estimated without the assumptions required for feature matching. The normal flow at a pixel is given by

\begin{equation}
n(p) = -\nabla_t I(p) \frac{\nabla_p I(p)}{\|\nabla_p I(p)\|^2}.
\end{equation}

The gradients are estimated using a Sobel filter with a kernel size of 3. To avoid erroneous estimation from gradients of small magnitude, pixels with gradients less than $15/255$ are discarded from the RMS normal flow magnitude calculation. The unstabilized frames experience flow magnitudes of approximately 1.1 pixels per frame. Stabilization reduces the normal flow magnitude to 0.5 to 0.9 pixels, however, it must be remembered that stabilization does not eliminate the measured rotational motion. Instead, it limits the high-frequency components of that motion. Finally, saccade-style stabilization reduces the normal flow magnitude to approximately 0.1 to 0.23 pixels.

Averaging frames results in a slight reduction in image sharpness. We measure image sharpness as the RMS magnitude of the image gradients estimated across all three color channels. While RMS contrast (or image variance) is a more typical measure of sharpness, it failed to quantify the reduction in sharpness due to frame averaging in this application. Our chosen measure indicates an approximately 10\% reduction in the sharpness measure when $N_{avg}=6$.

The average number of valid pixels, that is, pixels in $I^{\mathrm{stab}}$ which were filled by each of the averaged frames, is reported. The stabilized frames typically contain more than $99\%$ valid pixels when $N_{avg}=6$. Saccade stabilization achieves a lower score, as expected, but still typically above $97\%$.

We report the average angular velocity estimated by motion capture, the Flapper's onboard IMU, image based orientation estimate, and the stabilized view. As will be discussed later, the IMU and motion capture angular velocity estimates appear to be relatively inaccurate during flight, but accurate when the flapper is moved in-hand. We include them for completeness. The average angular velocity measured by the IMU is typically more than 250 deg/s and can be as high as 370 deg/s. Motion capture reports lower angular velocities ranging from 230 to about 300 deg/s. The image-based orientation estimator reports much lower angular velocities between 80 and 160 deg/s. Notably, in the in-hand sequence, where the flapper was not shaking, the IMU and image based angular velocity estimates are almost identical. The stabilized view's average angular velocity is between 30 and 130 deg/s in all sequences and typically below 90 deg/s. Note that the saccade stabilized view's angular velocity is zero at all times except when a saccade occurs. 

For the sequences recorded in the motion capture room, we report the flapper's average translational velocity which ranged between 0.9 and 1.7 m/s.

\subsection*{IMU Angular Velocity Estimates}

\begin{figure}[h]
     \centering
     \begin{subfigure}[t]{0.48\textwidth}
         \centering
         \includegraphics[width=\textwidth]{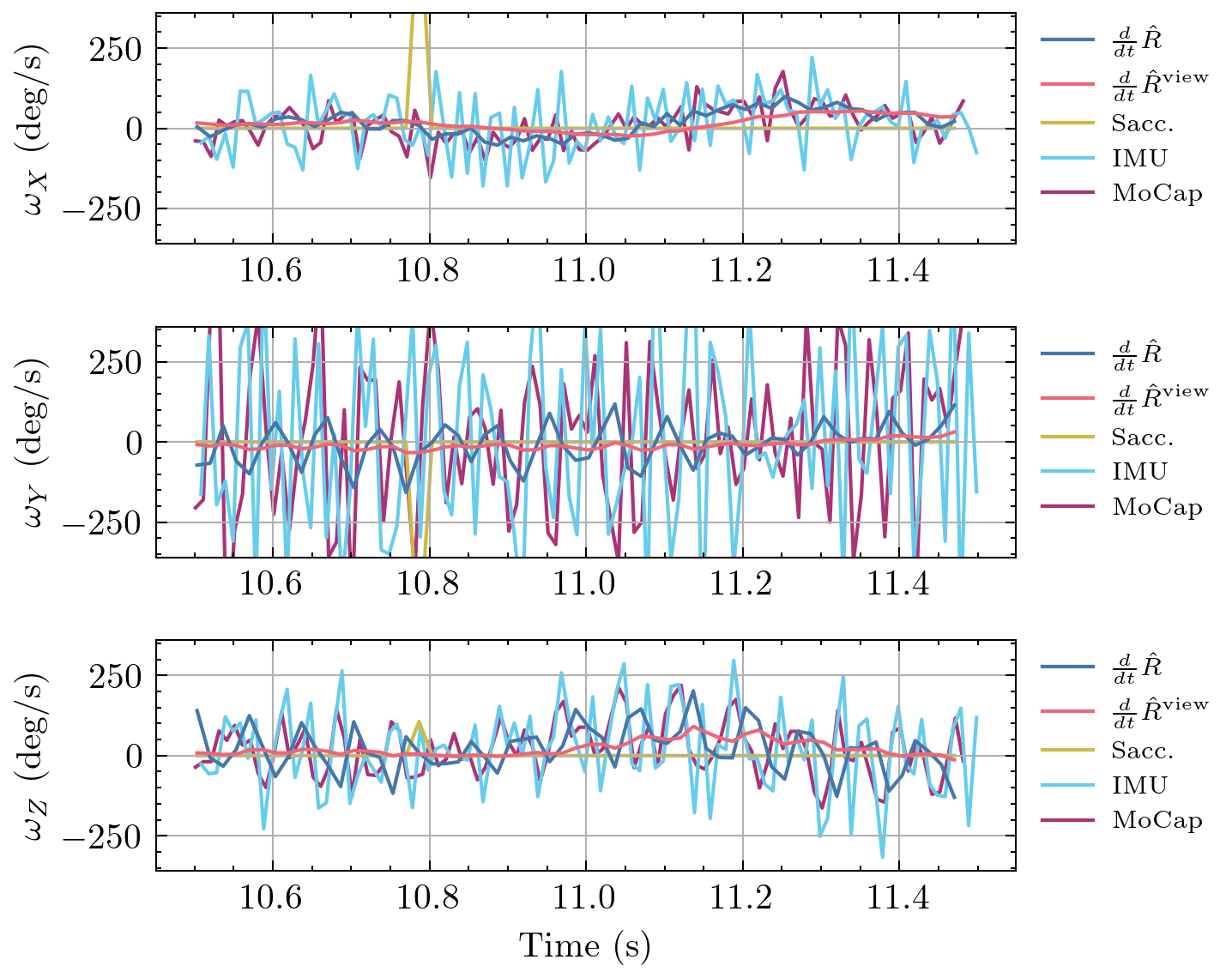}
         \label{fig:seq7imu}
     \end{subfigure}
     \hfill
     \begin{subfigure}[t]{0.48\textwidth}
         \centering
         \includegraphics[width=\textwidth]{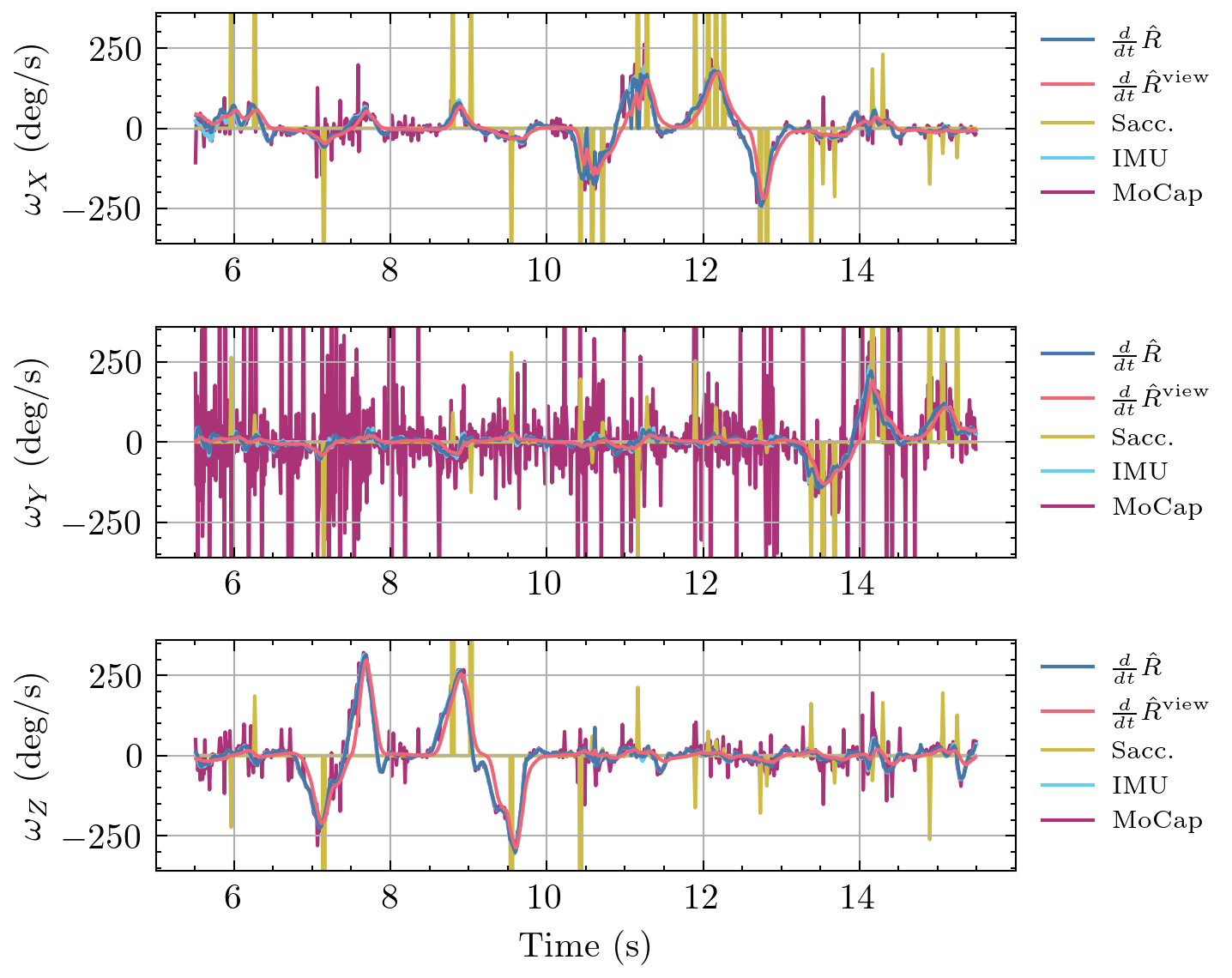}
         \label{fig:seq9imu}
     \end{subfigure}
        \caption{Zoomed in view of angular velocity estimates. \textbf{Left}: Angular velocity estimates from an onboard IMU, motion capture, the image orientation tracker, stabilized view, and saccade times during flight. The Flapper's onboard IMU fails to capture accurate enough angular velocity estimates for image stabilization during flight. \textbf{Right:} When moved smoothly by hand, the Flapper's onboard IMU's angular velocity estimates almost perfectly align with the image based angular velocity estimates. The motion capture system's angular velocity estimates also align, except for estimates along the Y axis, which are poor in quality because of occluded markers.}
        \label{fig:imucomp}
\end{figure}

The angular velocity estimates from the Flapper's onboard IMU are not accurate enough during flight to achieve an appreciable stabilization effect. This can be seen clearly from the left side of Figure \ref{fig:imucomp} where the IMU estimates are not aligned with image based velocity estimates computed from $\hat{R}$. In particular, the estimates from the IMU have a much larger magnitude than those measured from the image. However, when the Flapper is moved smoothly, as in the Hand sequence, the IMU's angular velocities almost perfectly match with our image-based estimate as shown on the right side of Figure \ref{fig:imucomp}. The motion capture system also has difficulty estimating the Flapper's angular velocity. However, this can be expected because the Flapper's body is semi-flexible, so the motion capture markers do not move rigidly with respect to each other.

\section*{Discussion}
Artificial Microsaccade Compensation produces the first stable, smooth videos from a tailless ornithopter. The algorithm runs in real time and relies on a carefully selected combination of video stabilization techniques that are suited for the platform. Further, it works with an inexpensive rolling shutter camera. Next, we contrast this specially designed stabilization algorithm with existing work on video stabilization.

Video stabilization is a well studied topic due to its numerous applications. Several distinct techniques have evolved \cite{wang2023videostab, Souza2022, guilluy2021}. Widely known techniques attempt to estimate the scene geometry and the camera's motion, then re-render the video to look as if it came from a camera that smoothly rotates and translates \cite{liu2009content}. Another popular technique is subspace methods, which split the image into many distinct parts that are stabilized separately, with some constraints on smoothness \cite{liu2011subspace}. Many methods result in gaps between separately stabilized regions of the image, which has led to the study of inpainting techniques that automatically fill those gaps \cite{matsushita2005fullframe, matsushita2006inpaint}. The predominant techniques rely heavily on feature matching, which requires significant effort to ensure feature trajectories are long enough and smooth enough to achieve satisfactory results \cite{lee2009robustfeatures, wang2013spattemp}. More recently, methods that incorporate the aforementioned concepts, but are based on deep learning and advanced representations such as Gaussian Splatting, have achieved impressive results but are typically far from achieving real-time performance and require significant training data \cite{guilluy2021, you2025gavs}.

Robotic applications require real-time, robust performance, and it is well known that feature matching across rapidly moving frames can fail because of motion blur. Further, our emphasis on microsaccade like motions means any estimated image motion should be constrained to that of a 3D rotation, which is difficult to enforce on feature trajectories. Stabilization based on 3D rotation and known intrinsics has been previously studied, but feature matching is still assumed \cite{chao2014}. Optical image stabilization based on angular velocity sensors have also been studied for some time \cite{sato1993}. However, applications typically focus on hand stabilization, which is relatively low bandwidth (\textless 10 Hz) \cite{la2015optical}. Deep learning based variants of angular velocity sensor based stabilization have also appeared \cite{liu2022DeepOIS}. In contrast, our desired application of stabilizing the video from a tailless ornithopter involves a base frequency of 12-20 Hz, which is augmented with higher order harmonics. Additionally, we find that the IMU on our tailless ornithopter does not accurately estimate the system's angular velocity during flight. Lastly, using gyroscopic sensors introduces additional complications, such as time synchronization and calibration of the gyroscopic sensor, which are often not well handled by low-cost hardware.

There are many directions for future work. Alternative methods for fusing frames, without reducing sharpness are of particular interest. Additionally, incorporating automatic identification of camera intrinsics would eliminate the need for prior calibration \cite{hartley1994rotcal}. Further, while the Flapper's IMU was not helpful, it is of interest to more carefully study the use of IMU's for rough compensation during aggressive shaking. Finally, applications to robot dogs and humanoids are of interest because they also shake when walking and running.

Determining the exact reason for the inaccuracy of the Flapper's IMU is the subject of future work. However, there are several possible explanations. The IMU is not mounted directly to the camera, instead it is on a nearby circuit board. Thus, the vibrations and semi-flexible body of the Flapper may cause the camera and IMU to move differently. Additionally, the Flapper's IMU may not be configured appropriately at the hardware level for measuring the Flapper's vibration patterns. Finally, it may be necessary to record the IMU at a rate higher than the maximum allowed by the Flapper's firmware, which is 100 Hz.
Regardless, the current results are acceptable for use in applications.

\section*{Methods}
\begin{figure}
  \centering
  \includegraphics[width=1.0\linewidth]{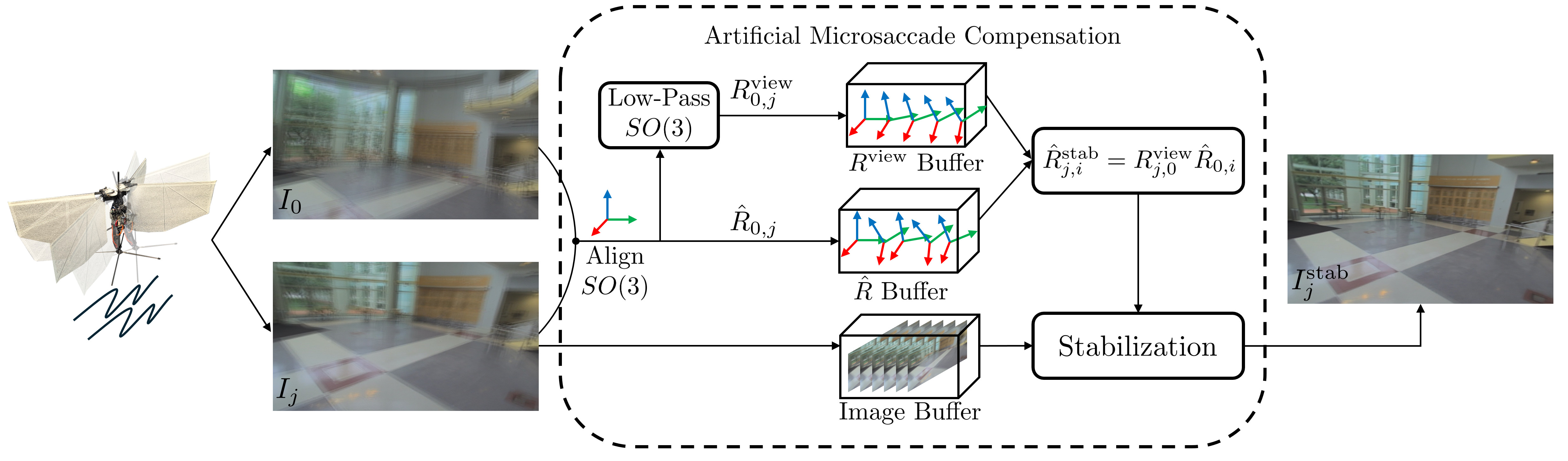}
  \caption{Artificial Microsaccade Compensation allows stabilizing unstable videos, such as those captured by the Flapper Nimble+, by directly matching incoming frames to a periodically updated template frame. Under the assumption of small rotational disturbances, a direct optimization of the mean squared error between images is used to continuously update an orientation estimate $\hat{R}$. Subsequently, a smoothed viewpoint $R^{\mathrm{view}}$ is computed. The simplest option for the stable viewpoint is a constant orientation. However, a more sophisticated approach that smoothly tracks the system's viewpoint can be realized with a low-pass filter on the group of rotation matrices $SO(3)$. The frames and rotations are buffered and used in the Artificial Microsaccade Compensation process, which combines multiple frames, taken at multiple times, to realize a high quality and stable video that is free from distortion.}
  \label{fig:algorithm}
\end{figure}

\begin{figure}
  \centering
  \includegraphics[width=1.0\linewidth]{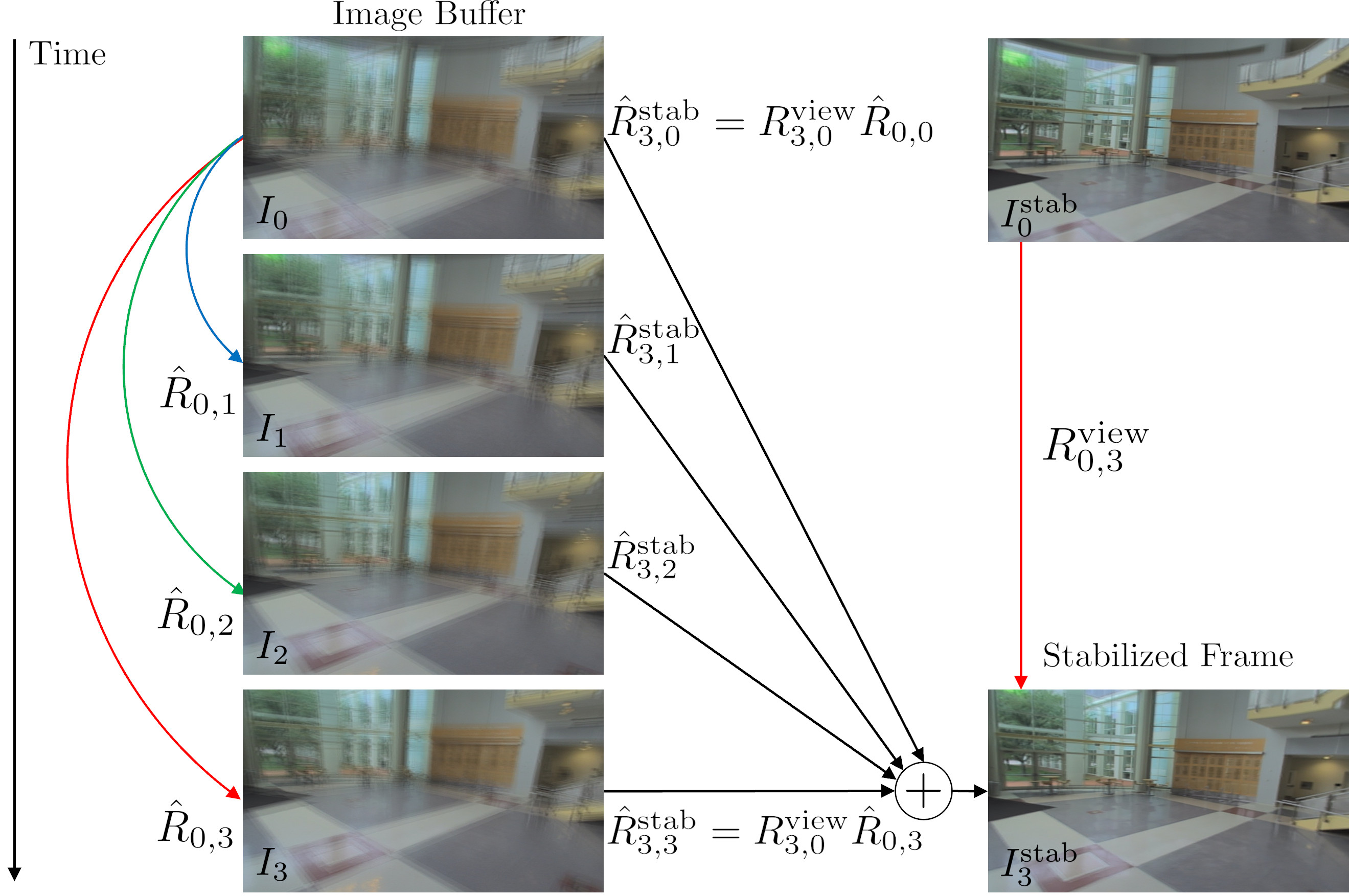}
  \caption{A detailed view of the ``Stabilization'' block in Figure \ref{fig:algorithm}. The left column is the average of 6 consecutive frames. While the individual frames are sharp, this averaging results in blur, which illustrates that the camera is shaking aggressively. The right column consists of the same six frames after stabilization and then averaging. In this paper, multiple frames are warped to the current stabilized view and averaged to reduce the effects of rolling shutter and transmission errors of a wireless camera onboard the Flapper. All displayed frames were computed using the camera's full resolution. Each timestep is associated with estimated orientations, $\hat{R}$ computed via a Lucas-Kanade tracker on the group of rotation matrices, $SO(3)$, and $R^{\mathrm{view}}$ a list of stable orientations computed by low-pass filtering $\hat{R}$. The stabilized output frame at the current time is computed by warping current and previous frames to the current stabilized viewpoint using stabilizing rotations $\hat{R}^{\mathrm{stab}}_{i,j} = R^{\mathrm{view}}_{i,0}\hat{R}_{0,j}$.}
  \label{fig:stabilization}
\end{figure}

The algorithm has four parts. First, we describe the direct method for aligning frames, which is an inverse Lucas-Kanade method that assumes a rotational warp function and known image intrinsics. Next, we describe the computation of stable view parameters via filtering on the group of rotations $SO(3)$. Next, we describe the stabilization procedure, which uses a buffer of previous frames and the computed orientations to suppress high-frequency rotations in the image. Finally, a variant based on purposeful saccades is described which results in an efficient recursive update. A block diagram is given in Figure \ref{fig:algorithm} while Figure \ref{fig:stabilization} provides a more detailed overview.

\subsection*{Rotation Estimation}
The Flapper has a mean flapping frequency of about 12 Hz, which causes the body and camera to oscillate with both translation and rotation at a similar frequency. However, the translational motion's contribution to optical flow is small in many applications. Thus, we propose ignoring the translational component of optical flow and only estimating the rotation under the assumption that the two frames are closely spaced in time. Formally, we assume the projection equations

\begin{equation}\label{eqn:projection}
p_c = \frac{1}{X^z_c} \begin{bmatrix}K & 0\end{bmatrix} T_{c,w} X_w,
\end{equation}

where $p_c$ is the pixel corresponding to the projection of point $X_w$ in the world frame into the camera frame, $X^z_c$ is the depth at pixel $p_c$, $T_{c,w}$ is the $SE(3)$ transform from world to camera frame, and $K$ is an invertible camera intrinsics matrix. In practice, the intrinsics of the fisheye lens used on the Flapper can be estimated beforehand. Thus, for the rest of the analysis, it is sufficient to assume $K = I$.

Then \eqref{eqn:projection} can be expanded to consider the rotational and translation components of $T_{c,w}$ as $R_{c,w}$ and $t_{c,w}$ respectively,

\begin{equation}\label{eqn:projectionexpanded}
p_c = \frac{1}{X^z_c} \left(R_{cw} X_w + t_{c,w}\right).
\end{equation}

If it is assumed that $t_{c,w}$ can be considered zero, then the projection can be manipulated to reveal a relationship that does not depend on depth. Let $r_{c,w}^3$ be the third row of $R_{c,w}$ so then $X^z_c = r_{c,w}^3 X_w$. Also, define $p_w = X_w / X_w^z$, that is, the projection of world point $X_w$ into the image when the camera is at and aligned with the origin. Then

\begin{equation}\label{eqn:projectionexpandednodepth}
\begin{split}
p_c &= \frac{R_{c,w} X_w}{r_{c,w}^3 X_w} = \frac{R_{cw} X_w / X_w^z}{r_{c,w}^3 X_w / X_w^z} = \frac{R_{c,w} p_w}{r_{cw}^3 p_w}. \\
\end{split}
\end{equation}

Here we see that a world point's projection into the image after a rotation depends only on its starting pixel projection and that rotation. Further, this relationship is easily inverted. Then, if we define two camera frames $c_k$ and $c_j$ we find

\begin{equation}
\begin{split}
p_{c_k} &= \frac{R_{c_k, w} p_w}{r_{c_k, w}^3 p_w} \\
p_{c_j} &= \frac{R_{c_j, w} p_w}{r_{c_j, w}^3 p_w} \\
\implies p_{c_k} &= \frac{R_{c_k, w} \left(r_{c_j, w}^3 p_w\right) R_{w, c_j} p_{c_j}}{r_{c_k, w}^3 \left(r_{c_j w}^3 p_w\right) R_{w, c_j} p_{c_j}} \\
&=  \frac{R_{c_k, c_j} p_{c_j}}{r_{c_k, c_j}^3 p_{c_j}}.
\end{split}\label{eqn:pixelcorrespond}
\end{equation}

Thus, the correspondences between the pixel coordinates in frame $c_k$ and $c_j$ are obtained using just the rotation $R_{c_k, c_j}$. Notably, this works only because translation was neglected. Otherwise, an estimate or model of depth would be required to compute $p_k$ from $p_j$. It is worth noting that homographies, also known as perspective warps, are commonly used to stabilize video feeds and generate panoramic images. However, homographies correspond to a planar depth assumption, which is well known to result in visible distortions when the scene is not planar. Thus, rotation is the only stabilizing effect that can be applied to an image without introducing distortion via erroneous assumption or estimation of depth.

In what follows, it can be assumed that all considered coordinate frames are a camera frame at some time. Thus, for brevity, the notation $R_{k,j}$ will be used to indicate the rotation from frame $c_j$ to $c_k$. Additionally, $p_k$ and $p_j$ will represent corresponding pixels in frames $k$ and $j$ taken from coordinate frames $c_k$ and $c_j$. Further, all pixel coordinates are considered homogenous so that $(R_{k,j} p_j) / (r_{k,j} p_j)$ can be written as $R_{k,j} p_j$. That is, the division by the third element of $R_{k,j} p_j$ is assumed.

Then, we can set up an optimization problem to estimate $R_{k,j}$ from $I_k$ and $I_j$ by warping $I_j$ so that the resulting pixel intensities are equal to those in $I_k$. The resulting problem is

\begin{equation}
\hat{R}_{k,j} = \argmin_{R_{k,j}\in SO(3)} \sum_{p_k \in I_k} \Big[I_k\big(p_k\big) - I_j \big(R_{j,k} p_k\big)\Big]^2.
\label{eqn:brightness}
\end{equation}

We solve this problem using an inverse compositional Lucas-Kanade method \cite{baker2004lucas}
specialized for a rotational warp given known camera intrinsics $K$. The inverse compositional formulation iterates over the following two steps

\begin{equation}
\begin{split}
\omega_{k,k-1} &= \argGN_{\omega\in so(3)} \sum_{p_k \in I_k} \Big[I_k\big(\exp\left(\omega\right) p_k \big) - I_j \big(R_{j,{k-1}} p_k\big)\Big]^2 \\
R_{j,k} &= R_{j,k-1} \exp\left(\omega_{k-1,k}\right)
\end{split}\label{eqn:brightnessinvcomp}
\end{equation}

where $\argGN$ returns the result of a single step of a Gauss-Newton minimization and $R_{j,k}$ is used as $R_{j,k-1}$ in the next iteration. $\exp$ denotes the matrix exponential, which maps elements of $so(3)$ to $SO(3)$. The inverse compositional formulation is essential because each iteration can reuse precomputed gradients of $I_k$ and the derivative of the rotational warp parameterized by the $so(3)$ Lie Algebra which determine the approximate Hessian used by the Gauss-Newton minimization.
These computations would otherwise be the most expensive computation of each iteration. To encourage stable tracking, we also incorporate a line search after each iteration. This guarantees that the loss decreases with each iteration, and prevents the estimate from diverging.

While the Lucas-Kanade method is well known, implementations of this variation are not widely available. Instead, most implementations consider the limited case of purely translational warps with coarse-to-fine refinement, as needed for tracking points. Additionally, some student implementations based on affine homographies are available, presumably because they are described in detail in \cite{baker2004lucas}. However, as previously noted, affine homographies and full homographies correspond to an incorrect assumption on depth. Further, they require estimating more parameters (six or eight, respectively), while our method requires only three. We include our implementation, which is based on Jax, with this paper's open-source code and the supplementary information.

It remains to track rotational changes across multiple frames. To do so, we propose the procedure described in Algorithm \ref{alg:tracking}. In brief, the procedure estimates rotational deltas, $\hat{R}_{k,j}$, using the current frame $I_j$ and a fixed template frame $I_k$ using the previously described Lucas-Kanade method. Each optimization is warmstarted with the previous rotation estimate $\hat{R}_{k,j-1}$. The first template frame is the first frame $I_0$. After $N_\mathrm{track}$ frames, it is assumed that $I_k$ should no longer be compared directly with the current frame $I_j$ because of the compounding effects of the unmodeled translation. At such times, the template frame $I_k$ is updated to be the current frame. Additionally, an estimate of the total rotation from time $0$ to the new time $k$, called $\hat{R}_{0,k}$, is updated using the value of $\hat{R}_{k,j}$ at the time of the template reset. The process repeats indefinitely.

\begin{algorithm}
\setstretch{1.1}
\caption{Artificial Microsaccade Compensation}\label{alg:tracking}
\begin{algorithmic}
\Require $N_{\mathrm{track}}, N_{\mathrm{avg}} \geq 0$
\State $I_k \gets$ get\_frame()
\State $\hat{R}_{0,k}, \hat{R}_{k,j}, R_{0,j}^{\mathrm{view}} \gets I$
\State $B_{I}, B_{\hat{R}}, B_{R^{\mathrm{view}}} \gets [ \quad ]$
\State $j \gets 0$
\While{True}
\\/* Update Orientation Estimate */
\State $I_j \gets$ get\_frame()
\State $\hat{R}_{k,j} \gets$ LK\_inv\_comp($I_k$, $I_j$, $\hat{R}_{k,j}$)
\State $\hat{R}_{0,j} \gets \hat{R}_{0,k} \hat{R}_{k,j}$
\\/* Update Stabilized View */
\State $\omega \gets \log\left(R_{j,0}^{\mathrm{view}} \hat{R}_{0,j}\right)$
\State $R_{0,j}^\mathrm{view} \gets R_{0,j}^\mathrm{view}\exp\left(\alpha\left(\|\omega\|\right)\omega \right)$
\\/* Append to Buffers */
\State $B_{I} \gets \mathrm{append}(B_{I}, I_j)$
\State $B_{\hat{R}} \gets \mathrm{append}(B_{\hat{R}} , \hat{R})$
\State $B_{R^{\mathrm{view}}} \gets \mathrm{append}(B_{R^{\mathrm{view}}}, R^{\mathrm{view}})$
\\/* Periodically Reset Template Image */
\If{$j \mod N_{\mathrm{track}} = N_{\mathrm{track}} - 1$}
    \State $I_k \gets I_j$
    \State $\hat{R}_{0,k} \gets \hat{R}_{0,j}$
    \State $\hat{R}_{k,j} \gets I$
\EndIf
\\/* Compute Stabilized Frame */
\State $B_{I^{\mathrm{stab}}} \gets [ \quad ]$
\For{$i \in \{j-N_{\mathrm{avg}},\ldots,j\}$}
\State $\hat{R}_{j,i}^{\mathrm{stab}} \gets B_{R^{\mathrm{view}}}[j]^T B_{\hat{R}}[i]$
\State $I^{\mathrm{stab}}_{j,i} \gets I^{\mathrm{stab}}_j + \mathrm{warp}(B_{I}[i], \hat{R}_{j,i}^{\mathrm{stab}})$
\State $B_{I^{\mathrm{stab}}}\gets \mathrm{append}(B_{I^{\mathrm{stab}}}, I^{\mathrm{stab}}_{j,i})$
\EndFor
\State $I^{\mathrm{stab}}_j \gets \mathrm{mean}\left(B_{I^{\mathrm{stab}}}\right)$
\State $j \gets j + 1$
\EndWhile
\end{algorithmic}
\end{algorithm}

Reusing the template frame $I_k$ for several iterations has several advantages. It limits drift in the cumulative orientation estimate because several new frames are compared to the same reference frame. Additionally, the expensive computation of the problem's gradients and approximate Hessian can be further postponed. While many criteria can be proposed to determine when to reset the template, such as the value of the residuals or the magnitude of the total rotation, we take the pragmatic approach of resetting every 5 frames.

\subsection*{Computation of a Stable View}
Next, we consider the estimation of a stable viewpoint. That is, a coordinate frame that follows the estimated orientation but does not shake. In unconstrained problems, a standard approach would be to use a low pass filter of the estimated parameters. However, in this case, the estimated parameters are on $SO(3)$ and so a low-pass filter on $SO(3)$ is needed. For completeness, we describe our methodology, but note that a similar approach appeared in \cite{chao2014}.

Let $R^{\mathrm{view}}_{0,j}$ be the chosen stable viewpoint and recall that $\hat{R}_{0,j}$ is the estimate of the current cumulative orientation. Then the rotation from the current orientation to the stable viewpoint is

\begin{equation}
R^{\mathrm{stab}}_{j,j} = R^{\mathrm{view}}_{j,0} \hat{R}_{0,j}
\end{equation}

because, by definition,

\begin{equation}
R^{\mathrm{view}}_{0,j} R^{\mathrm{stab}}_{j,j} = \hat{R}_{0,j}.
\end{equation}

If $R^{\mathrm{view}}_{0,j}$ follows $\hat{R}_{0,j}$, then $R^{\mathrm{stab}}_{j,j}$ will be close to $I$. Thus, it is safe to assume $w^{\mathrm{stab}}_{j,j} = \log(R^{\mathrm{stab}}_{j,j}) \in so(3)$ exists and represents the geodesic direction and magnitude from $R^{\mathrm{view}}_{0,j}$ to $\hat{R}_{0,j}$. This geodesic can then be used to interpolate between the two rotations according to

\begin{equation}
R^{\mathrm{interp}}(\beta) = R^{\mathrm{view}}_{0,j} \exp(\beta w^{\mathrm{stab}}_{j,j}), \quad \beta \in [0, 1].
\end{equation}

By definition, $R^{\mathrm{interp}}(0) = R^{\mathrm{view}}_{0,j}$ and $R^{\mathrm{interp}}(1) = \hat{R}_{0,j}$. Suppose, $\beta \in (0, 1)$, then a low pass filter on $SO(3)$ can be realized via the recursion

\begin{equation}
R^{\mathrm{view}}_{0,j+1} = R^{\mathrm{view}}_{0,j} \exp(\beta w^{\mathrm{stab}}_{j,j}).
\end{equation}

This filter is an analogue of a linear exponential filter. Such a filter may not converge fast enough to keep the stabilized view close to the true field of view, especially when the system is rotating quickly. We can improve the convergence at the expense of increasing the rotational noise by considering the magnitude of the rotation delta between the view and orientation estimate. Consider a monotonically increasing function $\alpha: \mathbb{R}\to\mathbb{R}$. Then the formulation can be generalized as

\begin{equation}
R^{\mathrm{view}}_{0,j+1} = R^{\mathrm{view}}_{0,j} \exp\left(\alpha\left(\|\omega^{\mathrm{stab}}_{j,j}\|\right)\omega^{\mathrm{stab}}_{j,j} \right)
\end{equation}

In practice, we find the $\alpha(\|\omega\|) = a + b\|\omega\|$ with $a = 2\Delta t$, $b = 40 \Delta t$, where $\Delta t$ is the interframe period provides reasonable results.

\subsection*{Stabilization}

At this point, the orientation of each frame has been estimated, $\hat{R}_{0,j}$ and a method for computing a stable viewpoint at each time, $R^{\mathrm{view}}_{0,j}$, has been proposed. It remains to estimate a stabilized video. Consider the rotation delta going from a frame's estimated orientation at time $i$ to any stable viewpoint at time $j$, we call this $R^{\mathrm{stab}}_{j,i}$, and it can be computed via

\begin{equation}
R^{\mathrm{stab}}_{j,i} = R^{\mathrm{view}}_{j,0} \hat{R}_{0,i}.
\end{equation}

With this rotation estimate, we can warp any previous frame so that it appears to have been taken from the orientation of any stabilized viewpoint. We refer to such a rotation as a stabilizing rotation. The stabilized frames can be computed via

\begin{equation}
I^{\mathrm{stab}}_{j,i}(p_j) = I_i(R^{\mathrm{stab}}_{i,j} p_j).
\end{equation}

If several such frames are computed for a particular time $j$, they can be combined, depending on the application, to produce a higher-quality stabilized frame. Specifically, in this paper, the stabilized frames are produced by averaging the last $N_{avg}=6$ frames after aligning them to the current stabilized view. This reduces rolling shutter effects and transmission artifacts introduced by the wireless camera mounted on the flapping robot. That is, the final stabilized image is given by

\begin{equation}
I^{\mathrm{stab}}_j = \frac{1}{N_{avg}} \sum_{i=j-N_{avg}}^{j} I^{\mathrm{stab}}_{j,i}
\label{eqn:computestable}
\end{equation}

\subsection*{Specialization to Saccades}
The previously described stabilization method is useful for human viewing because the stabilized frames appear as if they were captured by a camera rotating smoothly. However, it is also possible to remove all rotation from an image for short periods of time by forcing $R^{\mathrm{stab}}_{0,j}$ to be constant until it needs to be changed to maintain overlap between the stabilized and true field of views. This results in the least possible apparent image motion because the stabilized frames become as identical as possible, until the static orientation is updated. Further, the stabilized frame can be computed via a recursion which saves significant computation. Consider, if $R^{\mathrm{stab}}_{0,j} = R^{\mathrm{stab}}_{0,j+1}$ then we can drop the subscripts from these terms, because they are identical, resulting in the recursion

\begin{equation}
\begin{split}
I^{\mathrm{stab}}_{j+1}(p_j) = \frac{1}{N_{avg}} \sum_{i=j-N_{avg}+1}^{j+1}& I_i(\hat{R}_{i,0} R^{\mathrm{view}} p_j) \\
= \frac{1}{N_{avg}} \Big[I^{\mathrm{stab}}_{j}(p_j) &+ I_{j+1}(\hat{R}_{i,0} R^{\mathrm{view}} p_j) \\&- I_{j-N_{avg}}(\hat{R}_{i,0} R^{\mathrm{view}} p_j)\Big]. \\
\end{split}
\end{equation}

This expression adds and subtracts a frame from the previous stabilized frame to the get the next stabilized frame. In comparison, Equation \ref{eqn:computestable} requires summing $N_avg$ frames to produce a stabilized frame, which results in substantially more computation if $N_{avg} > 2$.

This recursion will produce suitable results until the true orientation of the camera rotates such that the true camera's field of view has little overlap with the stabilized image's field of view. At such times, the stabilized view must be shifted to align better with $\hat{R}_{0,j}$, similar to a saccade in human vision. For the purpose of study, when running in this recursive ``saccade'' mode, we reset $R^{\mathrm{stab}}$ to $\hat{R}_{0,j}$ when less than 90\% of $I^{\mathrm{stab}}$'s pixels are filled by each of the averaged frames.

\subsection*{Hardware}
The algorithm was tested on a Flapper Nimble+ manufactured by Flapper Drones. This tailless ornithopter beats its wings between approximately 12 and 20 Hz. The lack of a tail makes the platform agile, but also contributes to aggressive shaking. Our Flapper is equipped with a wireless HDZero Nano V3 FPV rolling shutter camera running at 1280$\times$720 resolution and 60 fps. The camera is popular among racing drone pilots.

\section*{Data Availability}
All data needed to interpret the conclusions of the paper are presented in the paper or supplementary information. Raw datasets generated and analyzed during the study are available at \url{prg.cs.umd.edu/AMC}.

\section*{Code Availability}
Code for the stabilization algorithm and analysis of raw data is available in the supplementary information and at \url{prg.cs.umd.edu/AMC}.

\section*{Acknowledgments}
The support of NSF award OISE 2020624 and the Maryland Robotics Center is gratefully acknowledged.

\section*{Author Contributions}
L.B. developed the algorithm, designed the study, conducted the experiments, and wrote the manuscript. G.C. helped designed the study, connected the results to the existing literature, and helped write the manuscript. Y.A. helped develop the algorithm and write the manuscript. All authors read and approved the final manuscript.

\section*{Competing Interests}
The authors declare no competing interests.



\bibliography{scifile}
\bibliographystyle{naturemag} 

\clearpage



\clearpage




\end{document}